\documentclass[conference]{IEEEtran}
\IEEEoverridecommandlockouts
\usepackage{cite}
\usepackage{amsmath,amssymb,amsfonts}

\usepackage{algorithm}
\usepackage{algpseudocode}

\usepackage{graphicx}
\usepackage{textcomp}
\usepackage{xcolor}

\usepackage{amsthm}

\usepackage{array}
\newcolumntype{?}{!{\vrule width 1.5pt}}

\usepackage{makecell}

\usepackage{hyperref}

\def\BibTeX{{\rm B\kern-.05em{\sc i\kern-.025em b}\kern-.08em
    T\kern-.1667em\lower.7ex\hbox{E}\kern-.125emX}}
\begin{document}

\title{Fast-Image2Point: Towards Real-Time Point Cloud Reconstruction of a Single Image using 3D Supervision \\
}

\makeatletter
\newcommand{\linebreakand}{%
  \end{@IEEEauthorhalign}
  \hfill\mbox{}\par
  \mbox{}\hfill\begin{@IEEEauthorhalign}
}
\makeatother

\author{
\IEEEauthorblockN{AmirHossein Zamani}
\IEEEauthorblockA{\textit{Electrical and Computer Engineering} \\
\textit{Concordia University}\\
Montreal, Canada \\
amirhossein.zamani@mail.concordia.ca}

\and

\IEEEauthorblockN{Amir G. Aghdam}
\IEEEauthorblockA{\textit{Electrical and Computer Engineering} \\
\textit{Concordia University}\\
Montreal, Canada \\
amir.aghdam@concordia.ca}

\and

\IEEEauthorblockN{Kamran Ghaffari T.}
\IEEEauthorblockA{\textit{Founder \& CEO} \\
\textit{Touché Technologies}\\
Montreal, Canada \\
kamran.ghaffari@touche-technologies.com}
}

\maketitle

\begin{abstract}
A key question in the problem of 3D reconstruction is how to train a machine or a robot to model 3D objects. Many tasks like navigation in real-time systems such as autonomous vehicles directly depend on this problem. These systems usually have limited computational power. Despite considerable progress in 3D reconstruction systems in recent years, applying them to real-time systems such as navigation systems in autonomous vehicles is still challenging due to the high complexity and computational demand of the existing methods. This study addresses current problems in reconstructing objects displayed in a single-view image in a faster (real-time) fashion. To this end, a simple yet powerful deep neural framework is developed. The proposed framework consists of two components: the feature extractor module and the 3D generator module. We use point cloud representation for the output of our reconstruction module. The ShapeNet dataset is utilized to compare the method with the existing results in terms of computation time and accuracy. Simulations demonstrate the superior performance of the proposed method. 
\end{abstract}


\section{Introduction}{

    3D reconstruction is a process of preparing geometric data of 3D objects with many applications in computer vision, computer graphics, and robotics \cite{RL}. With the rapid progress of virtual reality (VR) and augmented reality (AR) technologies in recent years, the demand for 3D data has significantly increased. Furthermore, real-world 3D data can also be used by real-time systems such as autonomous driving vehicles, unmanned aerial vehicles (UAV), and mobile robots for several tasks including localization, mapping, and navigation \cite{Indoor3D}. Most of these systems rely on the input data from a set of cameras to analyze the environment. This is because of the convenient image acquisition and its low cost compared to other sensory devices like Lidar. Having the assumption that the only sensory input is a single image, the problem of 3D reconstruction will be narrowed down into the 3D reconstruction of objects in a single image. Recovering lost information from just 2D images to model 3D objects is an ill-posed problem and potentially has infinite solutions. To tackle this problem, solutions based on multi-view image reconstruction are proposed. However, these solutions often require images captured from accurately calibrated cameras. The calibration is a complex process and is not feasible in most applications \cite{Survey1}. Thanks to the significantly growing volume of the available 3D data, data-driven approaches, especially deep learning-based methods, have led to the development of efficient techniques to reconstruct the 3D model of objects from one or multiple images. Using these techniques leads to removing the camera calibration process. Among these deep learning approaches, the supervised ones are widely used to construct 3D geometric data of an object from its RGB image using 2D or 3D supervision. These supervised methods utilize prior knowledge to formulate the 3D reconstruction problem as a recognition problem \cite{Survey1}.

    \subsection{Existing methods}
    {
        Object representation is an important factor to be considered in designing a neural framework for generating 3D shapes \cite{Pixel2Point}. In the computer vision literature, 3D objects are commonly represented in the format of voxel grids, octrees, 3D point clouds, 3D meshes, and implicit functions like SDF. Fig.~\ref{fig:3DRepresentation} shows this representation on the 3D model of a rabbit. Voxel grid is the least efficient format as it stores information about both occupied and unoccupied areas around the object, which is wasteful and leads to excessive computational load \cite{3DRep_Survey}. In addition, voxel grid representation suffers from blurring the fine details of the original 3D object \cite{multisingleimage}. Octrees are a structured hierarchy of varying-sized nested voxels. Although they are more efficient than the classic voxel grid, they exhibit poor performance for most real-time applications \cite{3DRep_Survey}. The 3D mesh representation constructs an object from a set of small, interconnected polygons (facets) described in terms of vertices in the 3D space. In AI applications, there are two major drawbacks to mesh representation. Firstly, the learning-based models based on mesh representation are not easily extendable. Second, processing large mesh datasets is computationally expensive and time-consuming for most real-time applications. Point clouds consist of unstructured data points scattered in a 3D space representing the surface of 3D objects \cite{Survey}. They are also easy to acquire as most depth-sensing 3D scanners collect large datasets of an object’s geometry in the format of point clouds \cite{Survey}. In AI applications, the unstructured nature of point cloud representation has computational advantages over other models.
        
        The first model which used point clouds for single-view 3D reconstruction was referred to as the point set generation network (PSGN) \cite{PSGN}. In that work, the authors experimented with different network design alternatives and two different loss functions: chamfer distance (CD) and earth mover’s distance (EMD). They proposed a simple vanilla architecture, a two-branch version, and an hourglass version. They outperformed the 3D-R2N2 \cite{3d-r2n2} architecture, a well-known work in the field of voxel reconstruction from images. 3D-LMNet proposed in \cite{3D-LMNet} is another model that operates by training a 3D point cloud auto-encoder and learns a function mapping from the 2D image space to the 3D point space. Similar to the PSGN, 3D-LMNet predicts multiple reconstruction outputs for a single-view input image to tackle uncertainty in the reconstruction process. DeformNet model \cite{DeformNet} addresses the same problem by performing an image-based shape retrieval and then applying a learned deformation to the retrieved shape template. More recently, Pixel2Point \cite{Pixel2Point} object reconstruction model has been proposed as a simple yet powerful autoencoder architecture. The model is trained in a 3D supervised fashion. Similar to our work, it generates a point cloud with a fixed number of points. An initial point cloud of a sphere shape is used to concatenate with the obtained features by an encoder to improve the quality of generated point cloud. In another recent work, the authors propose a network embedding three processes: image cropping, image retrieval, and point cloud reconstruction \cite{multisingleimage}. The image cropping module allows the network to isolate the desired object in multi-object images. The image retrieval module searches for similar images in the Shapenet dataset library \cite{Shapenet}. Then, ground truth point cloud and target images will be entered to the reconstruction network to obtain the 3D point cloud output. Most recently, the authors in \cite{VisualEnhanced1} challenge the chamfer distance and introduce a new loss function. They propose a supervised pipeline that focuses more on boundaries of a 3D point cloud object (edge and corner points). Despite significant progress in the accuracy and versatility of the existing methods in recent years, their ever-growing computational burden makes them impractical in most real-time applications.
        
    }
    \subsection{Core challenges}
    {
        The main objective of the aforementioned methods is to obtain a more accurate 3D point cloud of models. This effort often leads to designing more complex networks resulting in a higher computational load. In turn, a higher computational load not only slows down the network training process but also reduces the performance of the method in real-time applications. Hence, improving the accuracy of the method while maintaining its computational efficiency is a core challenge of real-time 3D reconstruction. Another common challenge is the lack of adequate data which is essential to training the network. This study intends to tackle the challenge of computational efficiency by proposing a novel Fast-Image2Point (FI2P) architecture.      
    }
    
    The rest of this article is organized as follows:
    Section 2 introduces the problem. Section 3 presents
    our method and implementation details. In section 4, we demonstrate and compare our qualitative and quantitative results with two existing methods in the literature. Finally, section 5 provides a conclusion and possible future directions in our work.

}
\section{Problem Introduction}
{

    \begin{figure}
        \centering
        \includegraphics[scale=0.07]{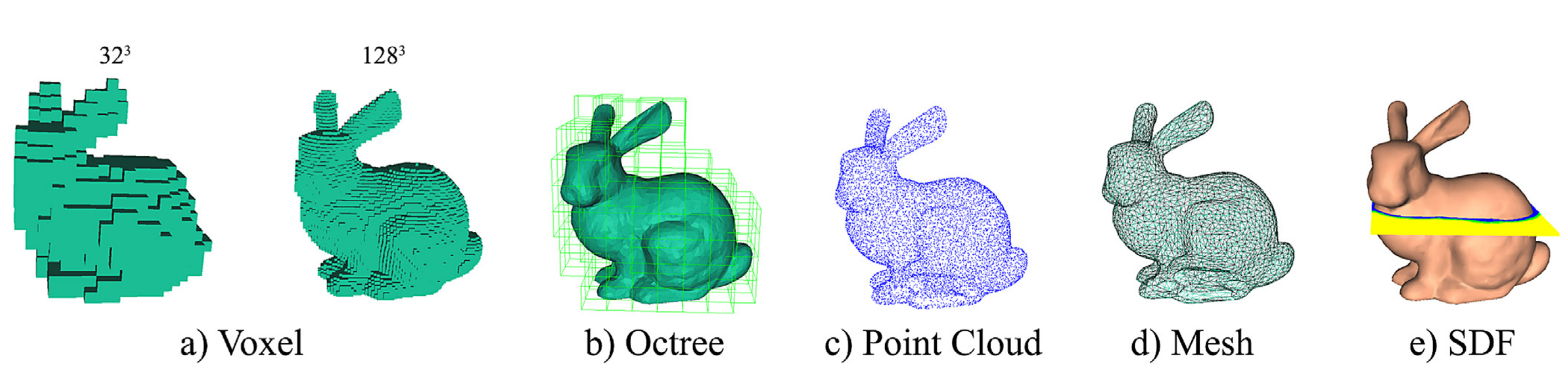}
        \caption{3D models of a rabbit shown by different 3D representations \cite{Survey}}
        \label{fig:3DRepresentation}
    \end{figure}

    3D reconstruction of an object from a single image involves processing the contents of that image in a specific way that results in an accurate estimate of the 3D geometrical data of the object. The accuracy of the 3D reconstruction method can be defined as the proximity of the estimated geometric data to the actual 3D geometry of the original object. Due to the generality and the inherent complexity of the 3D reconstruction problem, it is not feasible to design an analytical model that can associate the contents of an image of a random object to the object’s geometric data. The deep learning-based supervised technique is a suitable candidate to solve this category of unstructured problems.
    
    A properly designed deep learning supervised method can be trained to recognize and interpret the underlying patterns in a given image and provide an accurate estimate of the object’s geometry in the image. Such a training can be performed by using a large set of pre-identified data to systematically tune the network parameters until the desired accuracy is achieved.
    The contents of an image are stored as arrays of colored pixels with no specific correlation between them. Collectively, groups of these pixels represent different features of the captured object in the image. However, the number of constitutive pixels of an image is far larger than the fundamental features of that image. Such high-dimensional data can be excessively large to process in real-time. An effective solution to the high-dimensionality problem is to use a so-called encoder to extract the underlying features and patterns in the image and provide fewer but more representative datasets, to work with. A decoder can then be used to efficiently map the extracted features into the desired output, in this case, the 3D geometry data of the captured object in the image. This encoder-decoder pipeline is known as an autoencoder \cite{AutoEncoders}, \cite{AutoEncoders1}, \cite{AutoEncoders2} in the literature and has been used in a wider range of applications \cite{ApplicationAE}, \cite{ApplicationAE1}, \cite{ApplicationAE2}, including more recently, the image-based 3D geometric reconstruction \cite{PSGN}, \cite{3D-LMNet}, \cite{Pixel2Point}, \cite{VisualEnhanced1}, \cite{DensePCL}, \cite{3d-r2n2}, \cite{Atlas}. Fig.~\ref{fig:Autoencoder} shows a high-level representation of an autoencoder architecture.
    
    \begin{figure}
        \centering
        \includegraphics[scale=0.32]{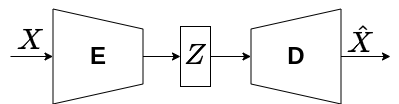}
        \caption{AutoEncoder architecture}
        \label{fig:Autoencoder}
    \end{figure}
    
    Accuracy and computational efficiency are the two most important target aspects of a real-time 3D reconstruction method. Aside from the quality of training, the accuracy of a supervised deep learning method highly depends on the constructional design of its network. On the other hand, the computational efficiency of the method also depends on the constructional design of its base network. Increasing the accuracy of 3D reconstruction has been the main focus of the existing methods, often at the cost of less computational efficiency which presents a bottleneck in real-time applications. The challenge that we aim to tackle in this work is designing a supervised deep learning network for real-time 3D reconstruction with sufficiently high accuracy and computational efficiency.
}

\section{Methodology}
{
    The FI2P method proposed in this work, applies the AutoEncoder technique \cite{AutoEncoders} to reconstruct the 3D geometric data of an object in 3D point cloud format from a single RGB image. Focusing on real-time applications, we design a computationally efficient network without compromising the accuracy of the method. This is done through a novel architecture of the encoder and decoder modules. The FI2P method is developed based on the following assumptions:
    
    \textit{(i)}{
        it only focuses on reconstructing the geometry of the captured object. The reconstruction of other qualities such as materials, lighting, view, and texture are out of the scope of this study. 
    }
    
    \textit{(ii)}{
        the input image is properly pre-processed using the existing image processing techniques to meet the FI2P’s requirements. The method requires the input image to have a white background and contain only a single object. The preprocessing techniques for image conditioning such as cropping, sharpening, masking, and balancing are readily available in the literature \cite{Preprocessing} and \cite{Preprocessing1}.
    }
    
    The following subsections elaborate the training procedure under the FI2P method.
    
    \subsection{Training Data} {
        As with most AI methods, FI2P network needs to be trained to properly interpret its input data. Training an AI network requires a large set of recognized high-quality data which can be challenging to obtain. FI2P training data, in particular, should contain the image of a known object alongside its corresponding 3D point cloud representation. Depending on the complexity of each category of objects, we suggest a minimum of 500 objects in the dataset to achieve acceptable accuracy. 
        To collect a large number of categorized 3D objects, we use ShapeNetCore \cite{Shapenet} libraries. This is a densely annotated subset of the ShapeNet dataset comprising 55 common object categories with 51,300 unique synthetic 3D models. These 3D models, however, were in mesh format and needed to be converted to the ground truth 3D point cloud.
        To convert the 3D models to point cloud and to capture a single image from them, we used the Open3D library \cite{Open3D}. The objects were positioned at the center of the scene and an image of each object was exported using a fixed-view camera for all objects. This way, we collectively prepared 7000 data pairs in 7 groups of datasets in categories of the airplane, bottle, car, bench, sofa, cellphone, and bike. The procedure is demonstrated in Fig.~\ref{fig:DatasetGeneration}.
        We dedicated a subdivision of the data in each category to the training (85\%), validation (5\%), and test (10\%) phases as explained in more detail in the next section.
        \begin{figure}
        \centering
        \includegraphics[scale=0.21]{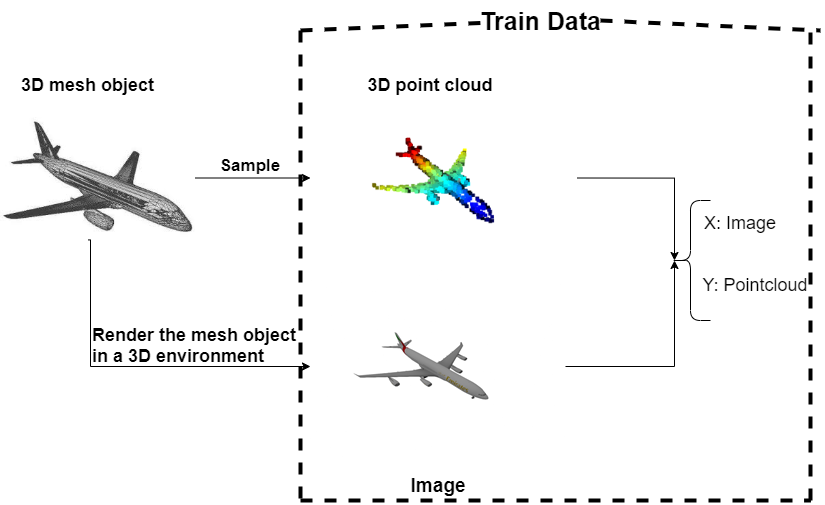}
        \caption{The dataset samples generation process for an airplane object}
        \label{fig:DatasetGeneration}
        \end{figure}
         
        \begin{table*}[htbp]
            \caption{An architectural comparison between the psgn model\cite{PSGN}, pixel2point model \cite{Pixel2Point}, and our proposed neural model (FI2P) \newline
            * The Pixel2Point model without the initial sphere \newline 
            ** The Pixel2Point model with an initial sphere with 16 points on its surface}
            
            \def\arraystretch{1}
                \begin{center}
                \begin{tabular}{|c?c|c|c|c|c|}
                \hline
                \textbf{Architecture element} & \textbf{PSGN \cite{PSGN}} & \textbf{Pixel2point (*) \cite{Pixel2Point}} & \textbf{Pixel2point (**) \cite{Pixel2Point}} & \textbf{Proposed method (FI2P)} \\ 
                \Xhline{2\arrayrulewidth}
                \textbf{Encoder} & ConvNet & ConvNet & ConvNet & ConvNet\\
                \hline
                \textbf{Decoder} & Deconv+FC (Parallel) & FC & FC & Deconv+FC (Sequential) \\
                \hline
                \textbf{Feature space size} & $512\times1\times1 = 512$ & $256\times1\times1=256$ & $16\times(3+256)=4144$ & $256\times8\times8=16384$\\
                \hline
                \textbf{Down-sampling operation} & Stride & Stride & Stride & Max-Pool+Stride\\
                \hline
        
                \end{tabular}
            \label{table:ArchitecturalComparison}
            \end{center}
        \end{table*}

    }
   
    \subsection{Network}{
    
        
        \begin{figure}
            \centering
            \includegraphics[scale=0.23]{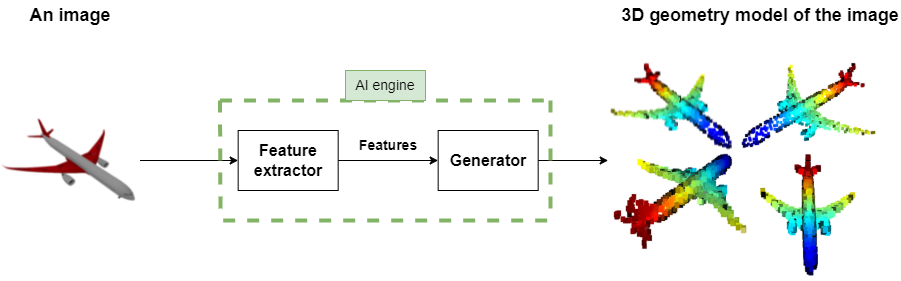}
            \caption{An overview of the FI2P system at the inference stage}
            \label{fig:TestTime}
        \end{figure}

        Our network uses 3D supervision, different from existing methods such as \cite{VisualEnhanced} and \cite{CAPNet}, to generate a 3D point cloud from a 2D single image. This category of methodologies use 2D supervision such as silhouettes, depth map, edge map, or corner map to model the 3D object. The proposed network performs two operations on the input image: encoding and decoding. In the first stage, the encoder extracts the most unique and salient features of the image. In the second stage, the decoder reconstructs the 3D point cloud model of the input image by generating the point cloud's positions $(x, y, z)$ based on the encoder's extracted features. The proposed network can be expressed by a composition function presented in equation (1):
        \begin{equation}
            \hat{X}=h_{d}(h_{e}(I, W_{e}), W_{d})
        \end{equation}
        where, $\hat{X}$ is the predicted 3D point cloud, $h_{d}$ is the decoder function, $h_{e}$ is the encoder function, $I$ is the input image, and $W_{e}$ and $W_{d}$ represent the parameters of the encoder and decoder (generator) functions, respectively. The $h_{e}$ and $h_{d}$ functions are obtained by designing and implementing two deep neural networks. To this end, we first design the high-level architecture of the network and then focus on the low-level architecture.
        
        \subsubsection{High-level design}{
                For the high-level design of the network, we considered three of its architectural aspects: neural layer type, non-linearity capability, and layer interconnection.
                
                (i) \textit{Neural layer type}:
                There are two types of neural layers commonly used in the literature: convolutional \cite{CNN} and fully-connected (FC) layers \cite{DeepLearningBook}. Convolutional layers are smaller than FC layers in terms of the number of parameters \cite{DeepLearningBook}. Table~\ref{table:CompareCNN_FC} illustrates a quantitative comparison between these two neural layers from the perspective of the number of parameters and computational operations for a fixed-size sample input image. Convolutional layers can also be implemented more efficiently on hardware chips. Several companies such as NVIDIA, Mobileye, Intel, Qualcomm, and Samsung actively develop CNN chips to enable real-time vision applications in smartphones, cameras, robots, and self-driving cars\cite{DeepLearning}. Thus, to increase the real-time capability of our network, we extensively used convolutional layers in our design.
            
            \begin{table*}[!h]
                \label{T:equipos}
                \caption{A quantitative comparison between convolution and fully-connected layers from the perspective of the number of parameters for the input image size of $320\times280$, kernel size of $2\times1$, and output size of $319\times280$ \cite{DeepLearningBook}
                }
                \centering
                
                \def\arraystretch{1}
        
                \begin{tabular}{|c|c|c|}  
                    \hline
                    \textbf{Params/Ops} & \textbf{Convolution} & \textbf{FC} \\
                    \cline{1-3}
                    \# of parameters & 2 & $319\times280\times320\times280 > 8e9$ \\ 
                    \# of operations & \textbf{$319\times280\times3 = 267960$} &  $> 16e9$ \\ 
                    
                    \hline
                \end{tabular}
                \label{table:CompareCNN_FC}
            \end{table*}

            (ii) \textit{Non-linearity capability}:
            Converting a 2D space (image) to a 3D space (point cloud) is a complex task due to its non-linear nature. We add non-linearity capability to our network by using two non-linear functions to effectively perform this conversion (mapping): (1) Rectified Linear Unit (ReLU) function as the activation function of hidden layers and (2) hyperbolic tangent (Tanh) function as the activation function of the last layer. The specific ReLU used in our design is a half-wave rectifier $f(z) = \max(z, 0)$ \cite{DeepLearning} and is used to add non-linearity capability to our reconstruction network in order to learn the complex mapping from 2D space (image) to 3D space (point cloud) more effectively. On the other hand, the Tanh function, is used to enforce the network's ability to construct the 3D point cloud model such that its points are mapped within the range of $[-1, 1]$. Note that the position value for each point in the ground truth point cloud is a floating-point number between -1 and 1. We observe that when we add the Tanh to our last layer, the network can be trained faster and learn a better solution than when there are no activation function for the last layer or there are other activation functions including ReLU, Sigmoid, and Softmax.   
            
            (iii) \textit{Layer interconnection}: By performing several test cases and implementing several deep neural architectures in the literature, we found out the interconnection among layers can directly affect the quality of the final solution for the 3D point cloud reconstruction task. PSGN\cite{PSGN} and Pixel2Point \cite{Pixel2Point} are two of several architectures we implemented for our test cases. In the PSGN, a parallel combination of convolution and FC layers is used in decoder design. In contrast, in the Pixel2Point network, a simple multi-layer perceptron (MLP) is used, containing several FC layers. In both architectures, we observe that using FC layers in the decoder immediately after the encoder causes the network to lose some spatial information in the input image. This is because FC layers require the processed image provided by the encoder to be converted to a vector, which leads to losing some spatial features that originally existed in the input image. To prevent such an issue, we use a cascaded architecture including convolution, deconvolution (Transpose operation of convolution), and two subsequent FC layers for our high-level architecture. According to our results, this sequential combination of layers reduces the computation time while preserving the reconstruction accuracy. We will further explain this point in the low-level design.

    }
        
    \subsubsection{Low-level design}{
        We focus on the elements that directly or indirectly affect computation time and accuracy in designing the encoder and decoder modules. This issue is discussed below.
    
        (i) \textit{Using higher-dimensional features}: Generally, we refer to the outputs of the encoder module as "features". These features represent the input image in a lower-dimensional space than the original space of the input image which is high-dimensional. The dimension of these features can directly affect the output quality of the downstream task to be used in the next layers of the network (specifically the "reconstruction" task in our case). The higher the dimension of the provided features by the encoder, the lower the probability of information loss during the encoding process. Therefore, we design the encoder such that the features are extracted in a higher-dimensional space compared to those in the \cite{PSGN} and \cite{Pixel2Point}, increasing the quality of encoder's output. We perform this by encoding the input image with five convolutional layers in our encoder. We use a relatively large feature space with the size of $256\times8\times8=16384$. In contrast, PSGN has a feature space size of $512\times1\times1=512$ and Pixel2Point has the feature space size of $256\times1\times1=256$ and $16\times(3+256)=4144$ in their two proposed networks.
        
        (ii) \textit{Extracting more spatial information}: Typically, there are extensive spatial correlations among pixels of an image. Extracting these spatial relationships and the underlying patterns in an image by a deep neural network is crucial for effectively reconstructing the 3D model of an object depicted in the input image. In this study, as mentioned before, a sequential architecture including convolution, deconvolution, and FC layers is used as the main skeleton for our network. This skeleton eliminates the need to convert the features provided by the encoder into a 1D array (a process referred to as \emph{flattening the feature space}). As a result, the features provided by our network preserve more spatial information about the object in the input image.
        
        (iii) \textit{Down-sampling operation}: During the encoding process, different down-sampling operations affect the computation time and accuracy of the network. Hence, to select a proper down-sampling function, we evaluate two of the most commonly used techniques in the literature: stride and max-pooling \cite{GuideToCNN}. To this end, we compared the effectiveness of these techniques by creating two versions of our network and analyze their effects on computation time and output loss. As shown in the Simulation Results section, the network with the stride approach achieves faster results while the one with the max-pool operator leads to more accurate results. To increase the real-time capability, we use the stride technique. Table~\ref{table:CompareOurs} summarizes these results for different categories of objects.
        
        Similarly, for designing the decoder, we consider the architectural aspects of the design described below:

        (i) \textit{Learning underlying point cloud patterns using deconvolution layers instead of FC layers}: The output of the encoder includes three types of features: high-level, mid-level, and low-level features. Respectively, these features contribute to the reconstruction process of high-level, mid-level, and low-level geometric features of the final 3D point cloud processed by the decoder. An ideal decoder would be able to fully harvest the embedded information contained within the provided features by the encoder to construct the base data from which the 3D point cloud can be constructed. Therefore, for this design stage, we select a layer that can more effectively preserve the fundamental image information to increase the construction accuracy at the next stage of the design.
        Subsequently, we consider two of the commonly used layers in the literature: Deconvolution and FC layers. As mentioned earlier, fully connected layers inherently removes the spatial information of the image. However, this is not the case for the deconvolution layers. Hence, we use deconvolution layers for this stage of design. 
        
        (ii) \textit{Mapping the extracted patterns to the final actual points in the output point cloud}: 
        To generate the point cloud, we need to explicitly assign each point $(x, y, z)$ to a neuron in our neural network. To do so, we utilize two FC layers without any usage of activation function to map the structure provided by the deconvolution layers in the previous step to explicit point cloud 3D points $(x, y, z)$. More specifically, the last FC layer contains $1024\times3 = 3072$ neurons, each representing the x, y, and z coordinates of a point in the point cloud. These two layers are the only FC layers that we use in our architecture. We also tested with one FC layer and two layers with ReLU activation functions after each layer, but the results were not favourable.

        \begin{figure} 
            \centering
            \includegraphics[scale=0.18]{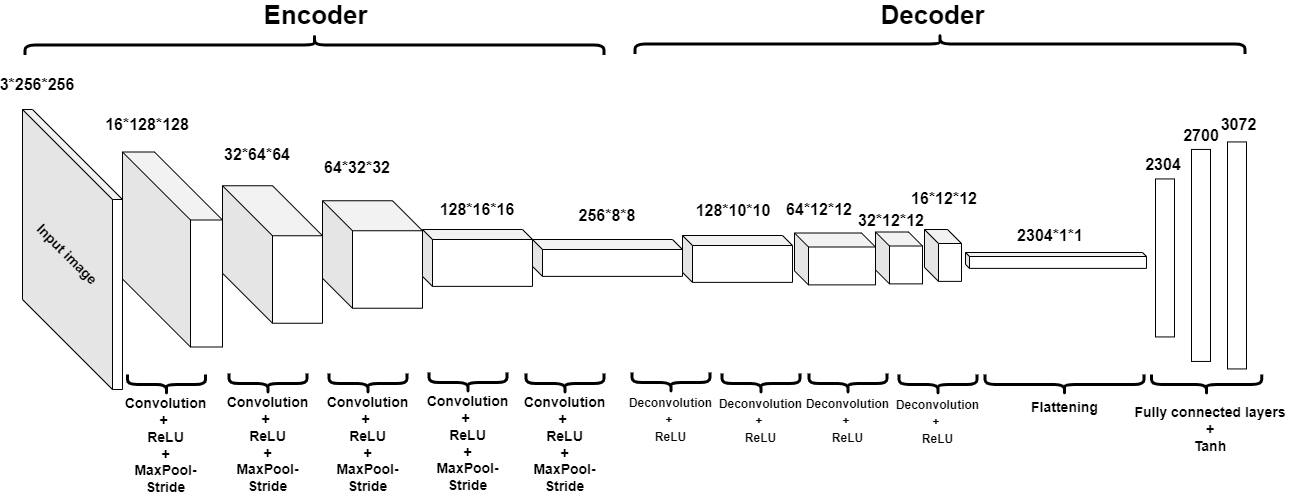}
            \caption{The architecture of 3D reconstruction neural network}
            \label{fig:Architecture}
        \end{figure}

    }
    
        Fig.~\ref{fig:TestTime} shows an overview of the network and the connectivity between the encoder and decoder modules at the test time. Fig.~\ref{fig:Architecture} illustrates the architecture of our network. This architecture is compared to the PSGN's \cite{PSGN} and the Pixel2Point's \cite{Pixel2Point} in Table~\ref{table:ArchitecturalComparison}. Also, Algorithm 1 presents the training procedure for our reconstruction network.

        \begin{algorithm}
            \caption{3D reconstruction network training}\label{alg:cap}
            \begin{algorithmic}
                \State $\eta \gets 0.0005$ \Comment{Initialize learning rate}
                \State $I \gets input Image$ \Comment{Get input image}
                \State $X \gets ground Truth$ \Comment{Get ground truth point cloud}
                \State $\epsilon \gets 0.001$ \Comment{Initialize $\epsilon$}
                \State $currentE \gets 0$ \Comment{Initialize current error}
                \State $lastE \gets 0$ \Comment{Initialize last error}
                \State $W \sim \mathcal{U}(-a, a)$ \Comment{Initialize weights uniformly (Xavier)}
                \While{$|currentE - lastE| < \epsilon$}
                    \State $lastE \gets currentE$
                    \State Pick batch of \textit{B} data points
                    \State $Z \gets h_{e}(I, W)$ \Comment{Extract features of the input image}
                    \State $\hat{X} \gets h_{d}(Z, W)$ \Comment{Predict the 3D point cloud}
                    \State $Loss \gets L_{k}(X, \hat{X})$ \Comment{Compute loss}
                    \State $\dfrac{\partial J(W)}{\partial W} \gets \dfrac{1}{B}\sum_{k=1}^{B} \dfrac{\partial Loss_{k}(W)}{\partial W}$ \Comment{Compute gradient}
                    \State $W \gets W - \eta \dfrac{\partial J(W)}{\partial W}$ \Comment{Update weights using the Adam optimizer}
                    \State $currentE \gets Loss$ \Comment{Update current error}
                    
                \EndWhile
            \end{algorithmic}
        \end{algorithm}
    }

    \subsection{Loss function}{
        In shape reconstruction from images using a supervised learning approach, we need to minimize the difference between the ground truth shape and the predicted 3D shape in the training process. A loss function is to be defined for this purpose which depends on the type of the 3D shape representation used in the training paradigm. We use point cloud as the 3D shape representation. Training shape reconstruction models can be done using either 3D supervision only \cite{PSGN}, \cite{Pixel2Point}, \cite{3D-LMNet}, \cite{3d-r2n2}, 2D supervision only \cite{VisualEnhanced}, \cite{CAPNet}, or a combination of both 2D and 3D supervision \cite{Survey}. In this work, for simplicity, we assume that we have 3D point clouds and we use 3D supervision to train the network. Hence, we need to define a loss function that measures the distance between the generated point cloud by the network and the ground truth point cloud. It should also be differentiable so that it can be used for updating weights of our neural architecture during the back-propagation step in the training procedure of our deep neural network \cite{DeepLearning}.
        
        In geometric representations, such as point cloud representations, the distance between the ground truth and predicted shapes can be measured by chamfer distance (CD) which was first introduced in \cite{PSGN} for 3D reconstruction. We also use this well-known measure to train our network. CD is defined as the squared distance between each point in one set to its nearest neighbor in the other set, is mathematically expressed as:
    
        \begin{equation}
            L(X, \hat{X})=\sum_{x \in X} \min _{\hat{x} \in \hat{X}}\|x-\hat{x}\|_{2}^{2}+\sum_{\hat{x} \in \hat{X}} \min _{x \in X}\|x-\hat{x}\|_{2}^{2}
        \end{equation}
        
        where $X$ and $\hat{X}$ are the ground truth point cloud set and predicted point cloud set, respectively. Moreover, $x$ and $\hat{x}$ are two corresponding points belonging to $X$ and $\hat{X}$, respectively.
    }
    
}

\section{Simulation Results}
{
    We implement, train, and evaluate the proposed method in Python using the Pytorch library \cite{Pytorch}. We also implemented and compare our supervised framework with other methods on seven data categories in the Shapenet dataset \cite{Shapenet}. For each object category, we train a separate model. Adam optimizer \cite{AdamOptimizer} is used to optimize the network parameters and weights with the initial learning rate of 0.00005, batch size of 5, and weight decay rate of 0.00001. We evaluate our proposed model from two perspectives: computation time and accuracy. We present the comparative results in three subsections in the sequel. First, we provide the general qualitative results of our proposed 3D reconstruction model on seven data categories of the ShapeNet dataset. Then, we quantitatively and qualitatively compare the results of the proposed model against PSGN \cite{PSGN} and Pixel2Point \cite{Pixel2Point}. Lastly, we analyze the effect of two different down-sampling techniques, pooling and stride, on the computation time of the models in the inference stage of the neural networks.

    \subsection{General results on the Shapenet dataset}{
        
        \begin{figure}
            \centering
            \includegraphics[scale=0.15]{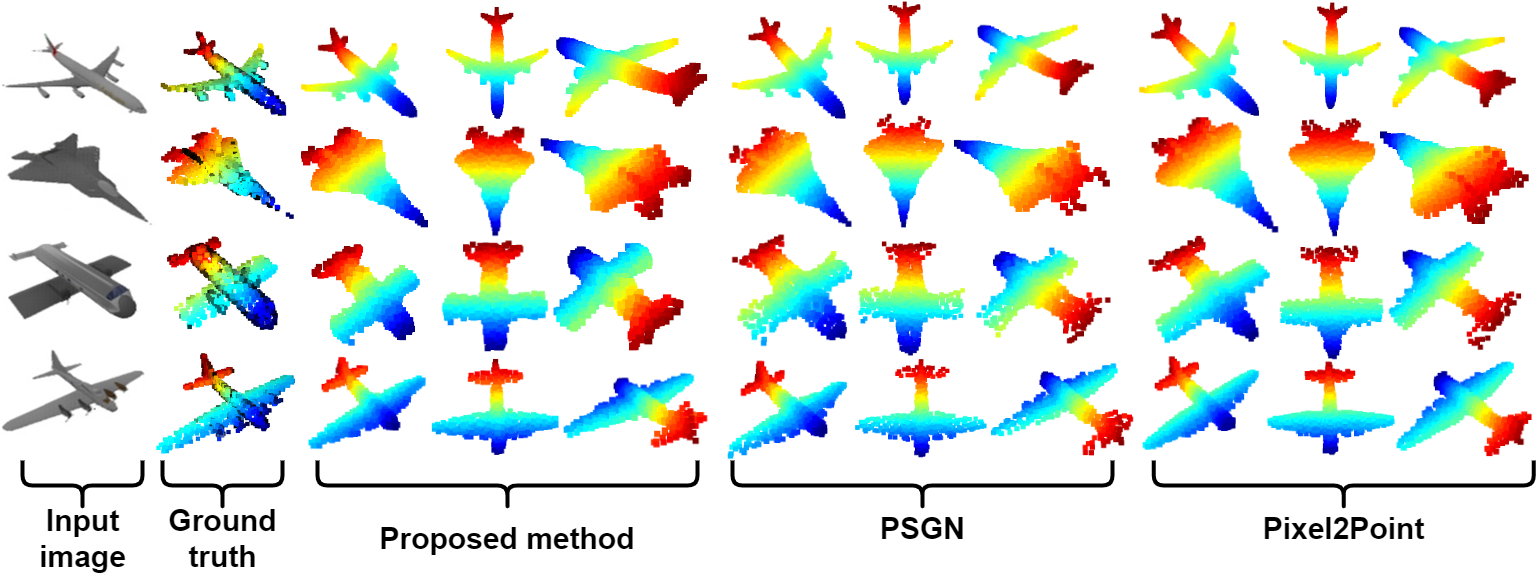}
            \caption{Qualitative 3D reconstruction results on the airplane dataset}
            \label{fig:AirplaneQLResult}
        \end{figure}
        
        
        \begin{figure}
            \centering
            \includegraphics[scale=0.16]{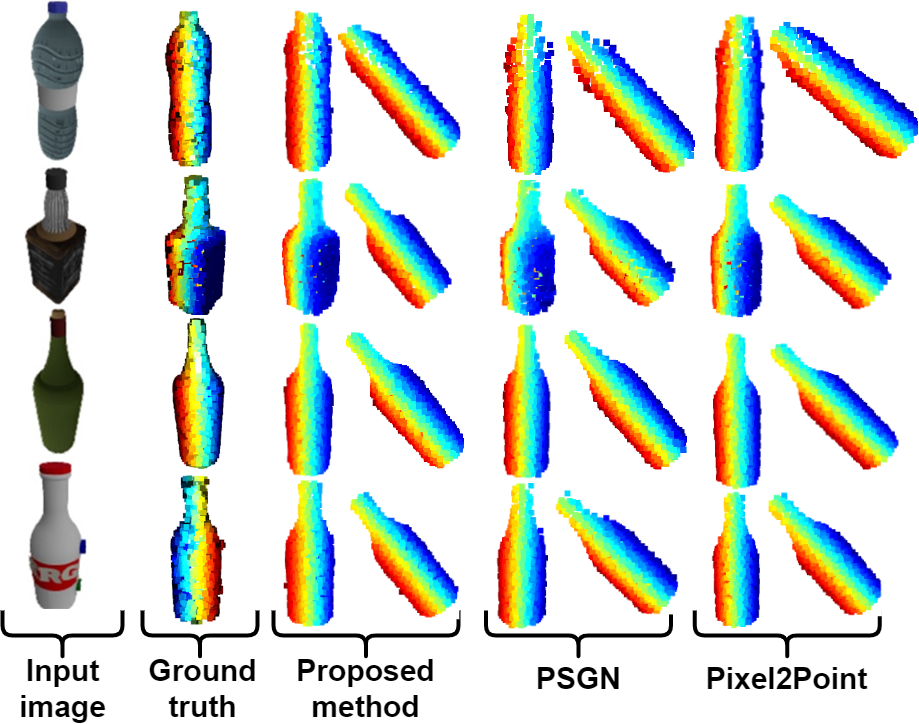}
            \caption{Qualitative 3D reconstruction results on the bottle dataset}
            \label{fig:BottleQLResult}
        \end{figure}
        
        The proposed model is trained, validated, and tested on the training, validation, and testing sets of the ShapeNet dataset. In the training stage, the model used synthetic images of 3D objects. They are rendered from one viewpoint and with constant lighting situations in the 3D rendering environment. After each training iteration, the model is applied to the unseen data (validation data) to validate its performance. To test the model, we use seven different data categories. We train and test the model for each data category separately. Figs.~\ref{fig:AirplaneQLResult} and \ref{fig:BottleQLResult} show the
        model's qualitative results on two categories of data. The results demonstrate that the system successfully recognized and generated the local and global structure of the object depicted in the input image.
    }
    
    \subsection{Comparison of two down-sampling techniques}{
        \begin{table}[!h]
            \label{T:equipos}
            \caption{A quantitative comparison between our two proposed models from the perspective of computation time and loss function value (lower is better)}
            \centering
            
            \def\arraystretch{1}
    
            \begin{tabular}{|c?cc?cc|}  
                \hline
                \textbf{Category} &
                \multicolumn{2}{c ?}{\textbf{Computation Time $(ms)$}} & \multicolumn{2}{c|}{\textbf{Chamfer Loss}} \\
                \cline{2-5}
                & Max-Pool & Stride & Max-Pool & Stride \\
                \hline
                Airplane    & 40.328 & \textbf{20.081} & 0.471 & \textbf{0.450} \\ 
                Bottle      & 44.556 & \textbf{17.517} & \textbf{1.248} & 1.354 \\ 
                Car         & 38.063 & \textbf{17.475} & \textbf{0.957} & 0.975 \\ 
                Sofa        & 40.959 & \textbf{19.921} & \textbf{1.755} & 1.780 \\ 
                Bench       & 45.661 & \textbf{17.584} & \textbf{1.301} & 1.438 \\ 
                Cellphone   & 49.283 & \textbf{17.932} & 1.340 & \textbf{1.297} \\ 
                Bike        & 39.969 & \textbf{17.340} & 1.831 & \textbf{1.765} \\ 
                \Xhline{1\arrayrulewidth}
                
                \textbf{Mean} & 42.688 & \textbf{18.264} & \textbf{1.272} & 1.294 \\ 
                \hline
            \end{tabular}
            \label{table:CompareOurs}
        \end{table}

        \begin{table*}[!h]
            \label{T:equipos}
            \caption{A quantitative comparison between psgn\cite{PSGN}, Pixel2Point \cite{Pixel2Point}, and our proposed method from the perspective of computation time and loss function value (lower is better) \newline 
            * The Pixel2Point model with an initial sphere with 16 points on its surface \newline 
            ** The Pixel2Point model without the initial sphere
            }
            \centering
            
            \def\arraystretch{1}
    
            \begin{tabular}{|c?cccc?cccc|}  
                \hline
                \textbf{Category} &
                \multicolumn{4}{c ?}{\textbf{Computation Time $(ms)$}} & \multicolumn{4}{c|}{\textbf{Chamfer Loss}} \\
                \cline{2-9}
                & FI2P & PSGN \cite{PSGN} & Pixel2Point (*) \cite{Pixel2Point} & Pixel2Point (**) \cite{Pixel2Point} & FI2P & PSGN \cite{PSGN} & Pixel2Point (*) \cite{Pixel2Point} & Pixel2Point (**) \cite{Pixel2Point} \\
                \hline
                Airplane    & \textbf{20.081} & 439.627 & 50.739 & 41.648 & \textbf{0.450} & 0.465 & 0.484 & 0.463      \\ 
                Bottle      & \textbf{17.517} & 445.045 & 63.831 & 38.414 & \textbf{1.354} & 1.482 & 1.514 & 1.379      \\ 
                Car         & \textbf{17.475} & 431.720 & 54.053 & 38.258 & \textbf{0.975} & 0.998 & 1.034 & 1.008      \\ 
                Sofa        & \textbf{19.921} & 420.135 & 57.779 & 41.264 & \textbf{1.780} & 1.833 & 1.869 & 1.879      \\ 
                Bench       & \textbf{17.584} & 419.151 & 45.532 & 37.027 & \textbf{1.438} & 2.170 & 1.516 & 1.444      \\ 
                Cellphone   & \textbf{17.932} & 405.803 & 49.454 & 34.390 & \textbf{1.297} & 1.353& 3.444 & 1.306       \\ 
                Bike   & \textbf{17.340} & 452.540 & 48.471 & 36.822 & \textbf{1.765} & 1.989 & 1.986 & 1.967        \\ 
                
                \Xhline{1\arrayrulewidth}
                
                \textbf{Mean} & \textbf{18.264} & 430.574 & 52.837 & 38.260 & \textbf{1.294} & 1.47 & 1.692 & 1.349 \\ \hline
            \end{tabular}
            \label{table:CompareOthers}
        \end{table*}

        We design and implement an autoencoder architecture for the task of 3D reconstruction. In the encoder part of this architecture, generally, there are two techniques to reduce the dimension of the input image (downsampling): pooling and strides \cite{DeepLearning}. Hence, we implemented our model with these two downsampling techniques and investigated their effect on the computation time and accuracy of the model. First, in the architecture of the encoder module, we use a max-pooling layer with the kernel size of 2 and strides 2 after each convolutional layer. As another test case, we implement the encoder part with strides of 2 (without any max-pool layers). We called these two test cases the \emph{Max-Pool} and \emph{Strides}, respectively, in Table~\ref{table:CompareOurs}. It can be observed from this table that the presence of the max-pooling layers among convolutional layers leads to longer computation time compared to the model using only strides as the downsampling technique. Figs.~\ref{fig:CompareOursLoss} and \ref{fig:CompareOurTime} shows the visual representation of data in Table~\ref{table:CompareOurs}.

    }

    \subsection{Comparison with other methods}{
    
        \begin{figure}
            \centering
            \includegraphics[scale=0.10]{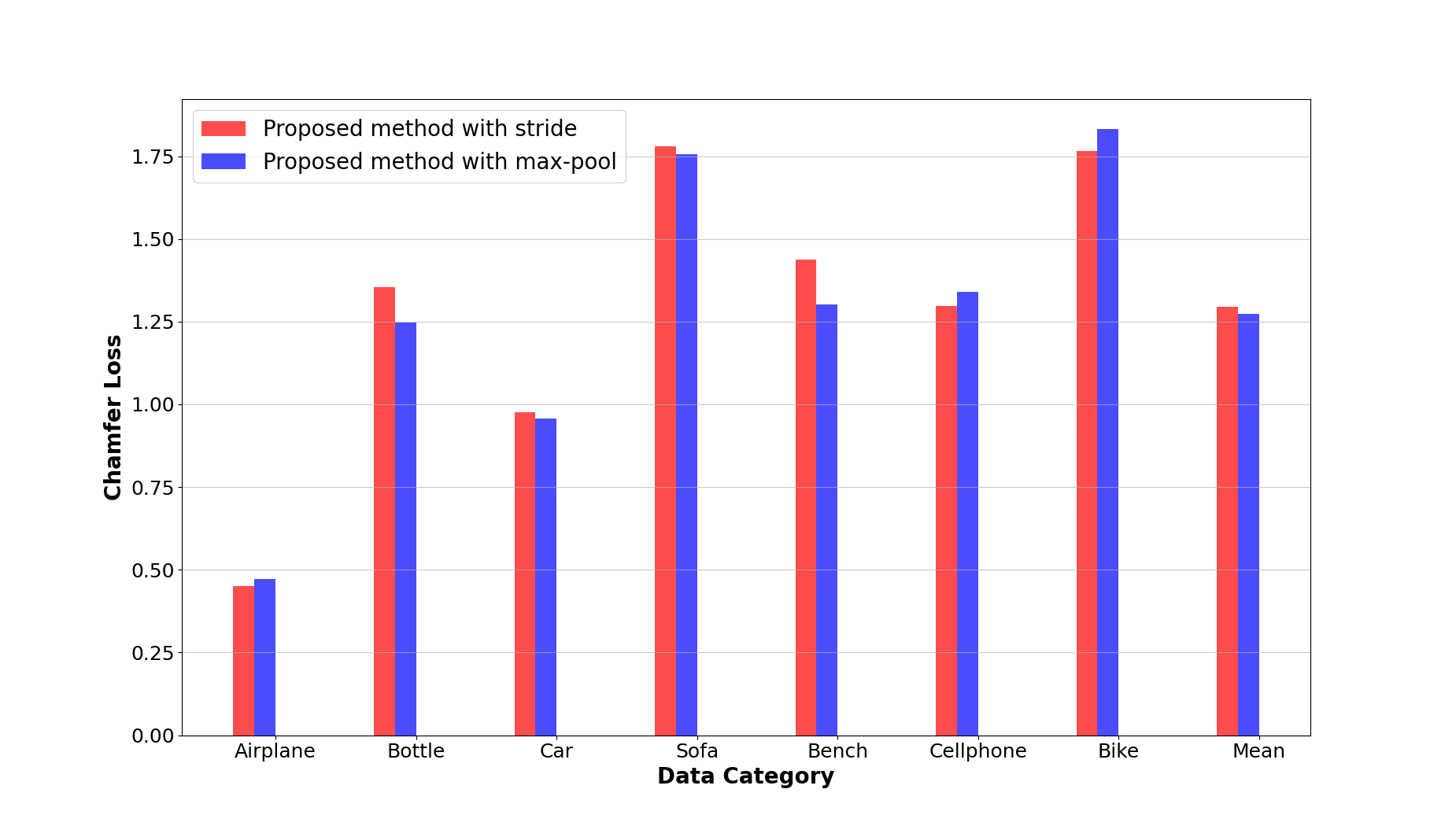}
            \caption{Visual comparison of Chamfer loss between the proposed model with maxpool and with stride downsampling operations}
            \label{fig:CompareOursLoss}
        \end{figure}
    
        \begin{figure}
            \centering
            \includegraphics[scale=0.105]{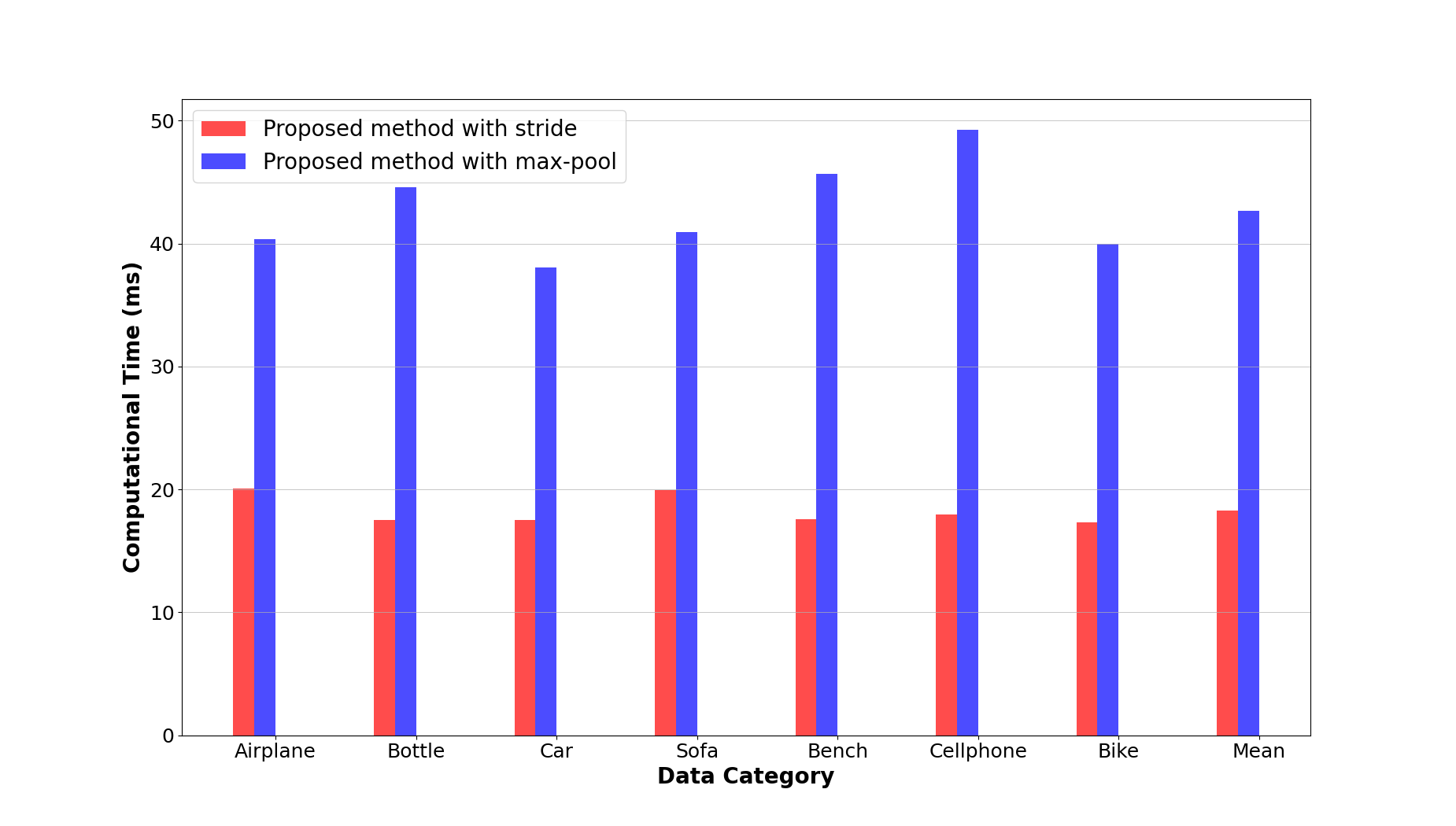}
            \caption{Visual comparison of computation time between the proposed model with maxpool and with stride downsampling operations}
            \label{fig:CompareOurTime}
        \end{figure}
        
        \begin{figure}
            \centering
            \includegraphics[scale=0.105]{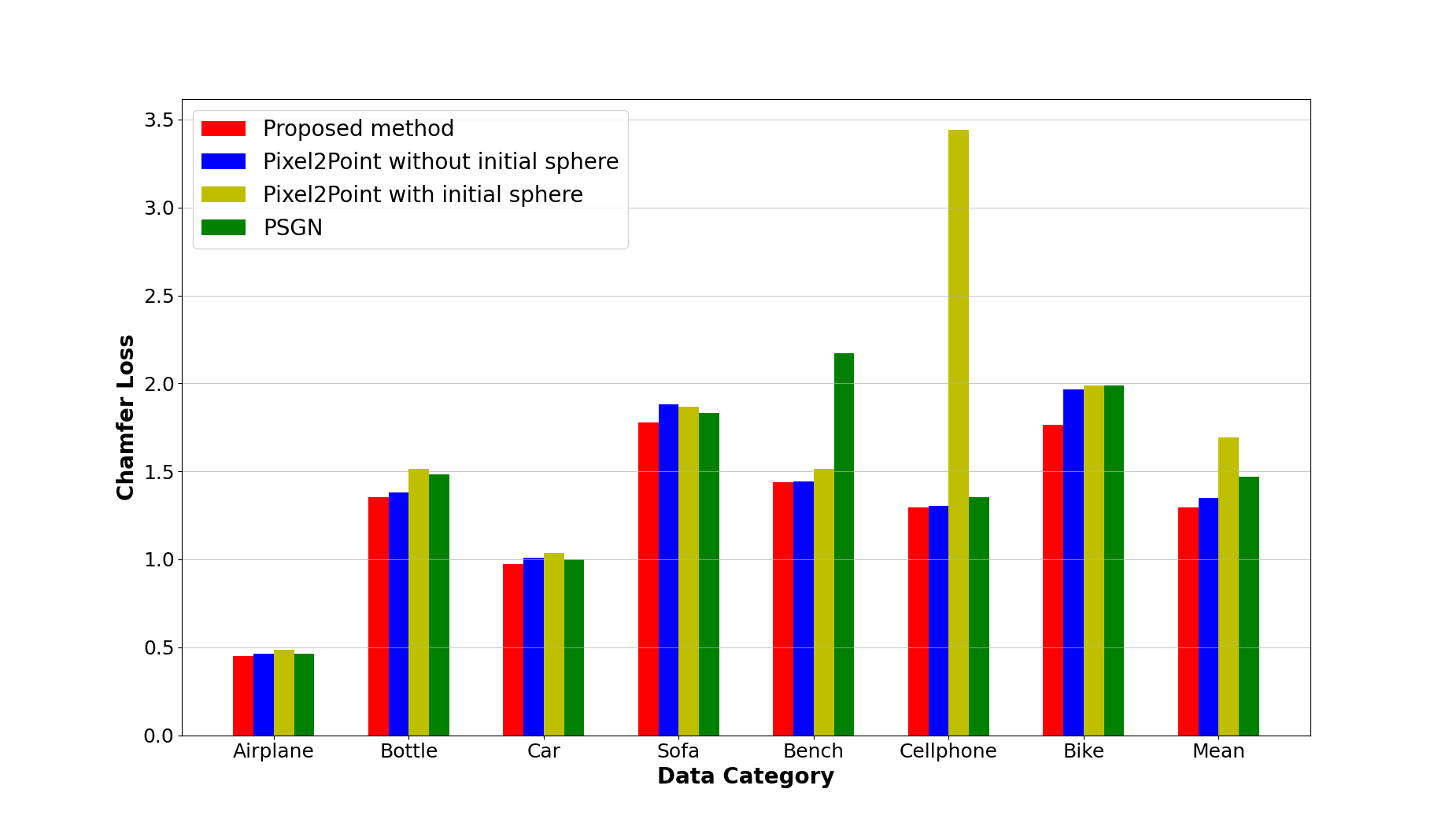}
            \caption{Visual comparison of Chamfer loss between the proposed method, PSGN \cite{PSGN}, and Pixel2Point \cite{Pixel2Point}}
        \end{figure}
        \begin{figure}
            \centering
            \includegraphics[scale=0.105]{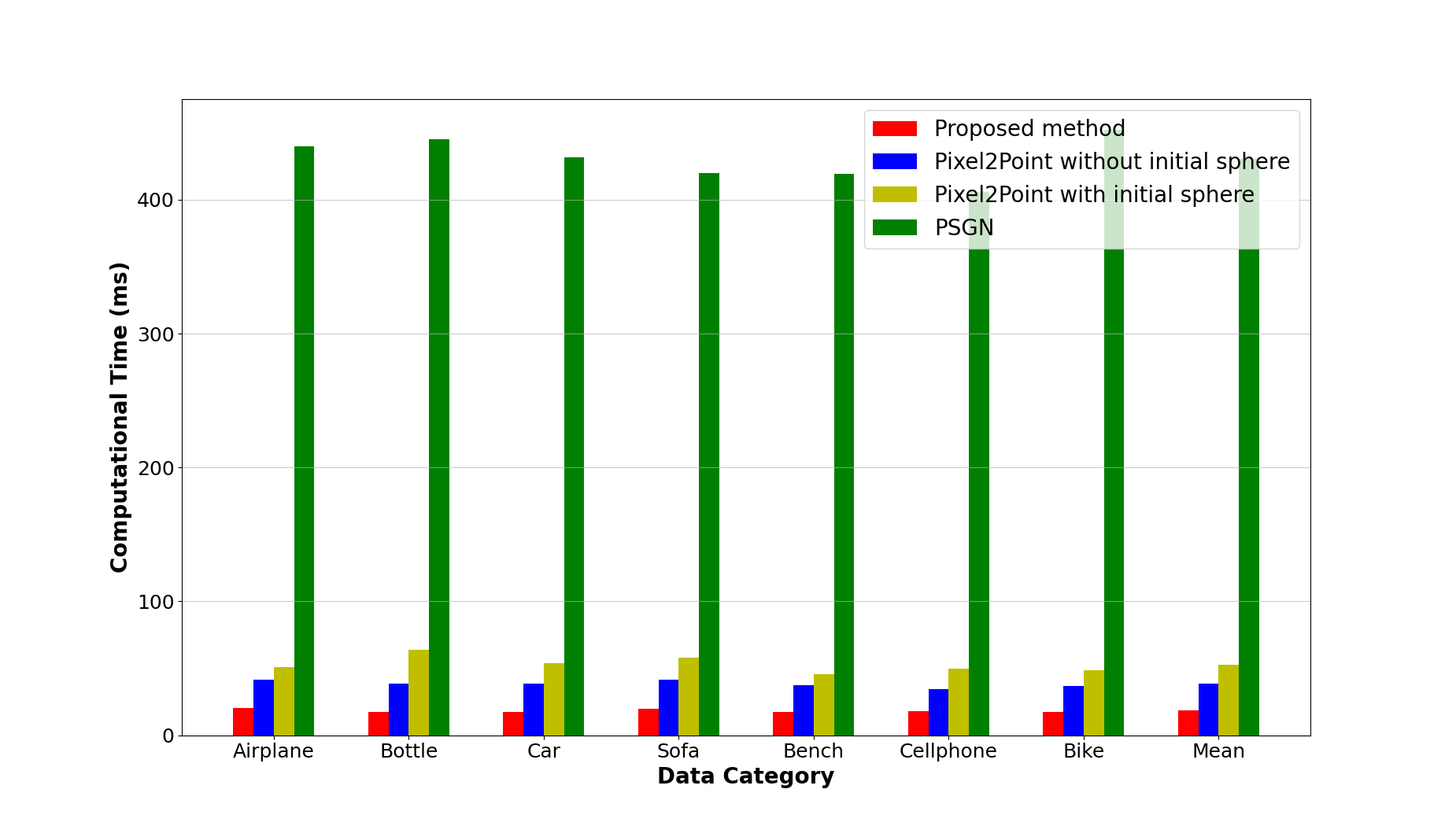}
            \caption{Visual comparison of computation time between the proposed method, PSGN \cite{PSGN}, and Pixel2Point \cite{Pixel2Point}}
        \end{figure}

        Figs.~\ref{fig:AirplaneQLResult} and \ref{fig:BottleQLResult} compare the proposed method with PSGN \cite{PSGN} and Pixel2Point \cite{Pixel2Point} for airplane and bottle data in terms of qualitative reconstruction. The first column in each figure is a $256\times256$ RGB image which is the input of our network. The second column is the ground truth data that the model aims to learn in order to generate the output similar to it. Other columns provide different views of the 3D reconstructed output of the network. We also quantitatively compare our results with the two models mentioned above. The result are summarized in Table~\ref{table:CompareOthers}. These results show that our network outperforms the ones proposed in \cite{PSGN} and \cite{Pixel2Point} from the viewpoint of computation time without losing accuracy. This advantage, however, comes at the cost of more storage requirements compared to Pixel2Point \cite{Pixel2Point}. Table~\ref{table:MemorySizeComparison} compares the three models in terms of memory usage. For a fair comparison, we train and test both methods under the same condition. We prepare the training data (2D images) with the size of $256\times256$ pixels from a single view and the label data (3D point clouds) with 1024 points.
        
        \begin{table}[htbp]
            \centering
            \caption{Memory consumption of psgn\cite{PSGN}, pixel2point \cite{Pixel2Point}, and the proposed method (lower is better)\newline 
            * The Pixel2Point model with an initial sphere with 16 points on its surface \newline 
            ** The Pixel2Point model without the initial sphere
            }
            
            \def\arraystretch{1}
                \begin{center}
                \begin{tabular}{|c|c|c|c|c|}
                \hline
                \textbf{} & \textbf{FI2P} & \textbf{PSGN\cite{PSGN}} & \textbf{P2P (*)\cite{Pixel2Point}} & \textbf{P2P (**) \cite{Pixel2Point}} \\ 
                \hline
                \textbf{Size} $(MB)$ & 61.2 & 119.7 & 269.5 & 15.0\\
                \hline
        
                \end{tabular}
            \label{table:MemorySizeComparison}
            \end{center}
        \end{table}

        We use 85\%, 5\%, and 10\% of the whole data for training, validation, and testing, respectively. Note that, we terminate the training process for each model when the smallest error for the validation data is reached, i.e., when there are no changes in the value of validation loss or when the validation loss starts to increase \cite{bishop}.

    }

}

\section{Conclusions and future work}{
    This work proposes a simple yet powerful auto-encoder architecture for the problem of real-time 3D reconstruction from a single RGB image. Using an encoder module, we first compress the 2D image to a unique featured vector. Then, we construct the 3D point cloud input object using the encoder module's obtained feature vector. Our network outperforms the ones proposed in \cite{PSGN} and \cite{Pixel2Point} from the viewpoint of computation time without losing accuracy. This advantage comes at the cost of more storage requirements compared to the Pixel2Point \cite{Pixel2Point}. However, memory usage is often not of concern in real-world applications. In our future work, we plan to improve the network architecture for more efficient storage requirements and lower computational complexity. We also aim to implement a real-time unsupervised learning approach for easier and faster training.
    
}

\section*{Acknowledgment}

This research has been supported by Touché Technologies and MITACS under the Accelerate Program.

{
    \bibliographystyle{./bibliography/IEEEtran}
    \bibliography{./bibliography/IEEEabrv,./bibliography/IEEEexample}

\begin{thebibliography}{10}
\providecommand{\url}[1]{#1}
\csname url@samestyle\endcsname
\providecommand{\newblock}{\relax}
\providecommand{\bibinfo}[2]{#2}
\providecommand{\BIBentrySTDinterwordspacing}{\spaceskip=0pt\relax}
\providecommand{\BIBentryALTinterwordstretchfactor}{4}
\providecommand{\BIBentryALTinterwordspacing}{\spaceskip=\fontdimen2\font plus
\BIBentryALTinterwordstretchfactor\fontdimen3\font minus
  \fontdimen4\font\relax}
\providecommand{\BIBforeignlanguage}[2]{{%
\expandafter\ifx\csname l@#1\endcsname\relax
\typeout{** WARNING: IEEEtran.bst: No hyphenation pattern has been}%
\typeout{** loaded for the language `#1'. Using the pattern for}%
\typeout{** the default language instead.}%
\else
\language=\csname l@#1\endcsname
\fi
#2}}
\providecommand{\BIBdecl}{\relax}
\BIBdecl

\bibitem{RL}
C.~Lin, T.~Fan, W.~Wang, and M.~Nie{\ss}ner, ``Modeling {3D} shapes by
  reinforcement learning,'' in \emph{European Conference on Computer
  Vision}.\hskip 1em plus 0.5em minus 0.4em\relax Springer, 2020, pp. 545--561.

\bibitem{Indoor3D}
M.~Yang and K.~Nagao, ``Automatic reconstruction of building-scale indoor {3D}
  environment with a deep-reinforcement-learning-based mobile robot,''
  \emph{International Journal of Robotics and Automation}, vol.~6, pp. 11--23,
  2019.

\bibitem{Survey1}
X.-F. Han, H.~Laga, and M.~Bennamoun, ``Image-based {3D} object reconstruction:
  State-of-the-art and trends in the deep learning era,'' \emph{IEEE
  transactions on pattern analysis and machine intelligence}, vol.~43, no.~5,
  pp. 1578--1604, 2019.

\bibitem{Pixel2Point}
A.~J. Afifi, J.~Magnusson, T.~A. Soomro, and O.~Hellwich, ``Pixel2point: {3D}
  object reconstruction from a single image using cnn and initial sphere,''
  \emph{IEEE Access}, vol.~9, pp. 110--121, 2020.

\bibitem{3DRep_Survey}
E.~Ahmed, A.~Saint, A.~Shabayek, K.~Cherenkova, R.~Das, G.~Gusev, D.~Aouada,
  and B.~Ottersten, ``Deep learning advances on different {3D} data
  representations: A survey. ar{X}iv 2018,'' \emph{arXiv preprint
  arXiv:1808.01462}.

\bibitem{multisingleimage}
H.~Chen, Y.~Zuo, Y.~Tong, and L.~Zhu, ``{3D} point cloud generation
  reconstruction from single image based on image retrieval,'' \emph{Results in
  Optics}, vol.~5, p. 100124, 2021.

\bibitem{Survey}
G.~Fahim, K.~Amin, and S.~Zarif, ``Single-view {3D} reconstruction: a survey of
  deep learning methods,'' \emph{Computers \& Graphics}, vol.~94, pp. 164--190,
  2021.

\bibitem{PSGN}
H.~Fan, H.~Su, and L.~J. Guibas, ``A point set generation network for {3D}
  object reconstruction from a single image,'' in \emph{Proceedings of the IEEE
  Conference on Computer Vision and Pattern Recognition}, 2017, pp. 605--613.

\bibitem{3d-r2n2}
C.~B. Choy, D.~Xu, J.~Gwak, K.~Chen, and S.~Savarese, ``3d-r2n2: A unified
  approach for single and multi-view {3D} object reconstruction,'' in
  \emph{European Conference on Computer Vision}.\hskip 1em plus 0.5em minus
  0.4em\relax Springer, 2016, pp. 628--644.

\bibitem{3D-LMNet}
P.~Mandikal, K.~Navaneet, M.~Agarwal, and R.~V. Babu, ``{3D-LMN}et: Latent
  embedding matching for accurate and diverse {3D} point cloud reconstruction
  from a single image,'' \emph{arXiv preprint arXiv:1807.07796}, 2018.

\bibitem{DeformNet}
A.~Kurenkov, J.~Ji, A.~Garg, V.~Mehta, J.~Gwak, C.~Choy, and S.~Savarese,
  ``Deformnet: Free-form deformation network for {3D} shape reconstruction from
  a single image,'' in \emph{2018 IEEE Winter Conference on Applications of
  Computer Vision (WACV)}.\hskip 1em plus 0.5em minus 0.4em\relax IEEE, 2018,
  pp. 858--866.

\bibitem{Shapenet}
A.~X. Chang, T.~Funkhouser, L.~Guibas, P.~Hanrahan, Q.~Huang, Z.~Li,
  S.~Savarese, M.~Savva, S.~Song, H.~Su \emph{et~al.}, ``Shapenet: An
  information-rich {3D} model repository,'' \emph{arXiv preprint
  arXiv:1512.03012}, 2015.

\bibitem{VisualEnhanced1}
G.~Ping, M.~A. Esfahani, J.~Chen, and H.~Wang, ``Visual enhancement of
  single-view {3D} point cloud reconstruction,'' \emph{Computers \& Graphics},
  2022.

\bibitem{AutoEncoders}
G.~E. Hinton and R.~R. Salakhutdinov, ``Reducing the dimensionality of data
  with neural networks,'' \emph{science}, vol. 313, no. 5786, pp. 504--507,
  2006.

\bibitem{AutoEncoders1}
P.~Baldi, ``Autoencoders, unsupervised learning, and deep architectures,'' in
  \emph{Proceedings of ICML workshop on unsupervised and transfer
  learning}.\hskip 1em plus 0.5em minus 0.4em\relax JMLR Workshop and
  Conference Proceedings, 2012, pp. 37--49.

\bibitem{AutoEncoders2}
W.~Wang, Y.~Huang, Y.~Wang, and L.~Wang, ``Generalized autoencoder: A neural
  network framework for dimensionality reduction,'' in \emph{Proceedings of the
  IEEE conference on computer vision and pattern recognition workshops}, 2014,
  pp. 490--497.

\bibitem{ApplicationAE}
T.~Blaschke, M.~Olivecrona, O.~Engkvist, J.~Bajorath, and H.~Chen,
  ``Application of generative autoencoder in de novo molecular design,''
  \emph{Molecular informatics}, vol.~37, no. 1-2, p. 1700123, 2018.

\bibitem{ApplicationAE1}
Q.~Zhao, E.~Adeli, N.~Honnorat, T.~Leng, and K.~M. Pohl, ``Variational
  autoencoder for regression: Application to brain aging analysis,'' in
  \emph{International Conference on Medical Image Computing and
  Computer-Assisted Intervention}.\hskip 1em plus 0.5em minus 0.4em\relax
  Springer, 2019, pp. 823--831.

\bibitem{ApplicationAE2}
T.~R. Phillips, C.~E. Heaney, P.~N. Smith, and C.~C. Pain, ``An
  autoencoder-based reduced-order model for eigenvalue problems with
  application to neutron diffusion,'' \emph{International Journal for Numerical
  Methods in Engineering}, vol. 122, no.~15, pp. 3780--3811, 2021.

\bibitem{DensePCL}
T.~Hu, G.~Lin, Z.~Han, and M.~Zwicker, ``Learning to generate dense point
  clouds with textures on multiple categories,'' in \emph{Proceedings of the
  IEEE/CVF Winter Conference on Applications of Computer Vision}, 2021, pp.
  2170--2179.

\bibitem{Atlas}
Z.~Murez, T.~v. As, J.~Bartolozzi, A.~Sinha, V.~Badrinarayanan, and
  A.~Rabinovich, ``Atlas: End-to-end {3D} scene reconstruction from posed
  images,'' in \emph{European Conference on Computer Vision}.\hskip 1em plus
  0.5em minus 0.4em\relax Springer, 2020, pp. 414--431.

\bibitem{Preprocessing}
S.~Bhattacharyya, ``A brief survey of color image preprocessing and
  segmentation techniques,'' \emph{Journal of Pattern Recognition Research},
  vol.~1, no.~1, pp. 120--129, 2011.

\bibitem{Preprocessing1}
K.~K. Pal and K.~Sudeep, ``Preprocessing for image classification by
  convolutional neural networks,'' in \emph{2016 IEEE International Conference
  on Recent Trends in Electronics, Information \& Communication Technology
  (RTEICT)}.\hskip 1em plus 0.5em minus 0.4em\relax IEEE, 2016, pp. 1778--1781.

\bibitem{Open3D}
Q.-Y. Zhou, J.~Park, and V.~Koltun, ``{Open3D}: {A} modern library for {3D}
  data processing,'' \emph{arXiv:1801.09847}, 2018.

\bibitem{VisualEnhanced}
G.~Ping, M.~A. Esfahani, and H.~Wang, ``Visual enhanced {3D} point cloud
  reconstruction from a single image,'' \emph{arXiv preprint arXiv:2108.07685},
  2021.

\bibitem{CAPNet}
K.~Navaneet, P.~Mandikal, M.~Agarwal, and R.~V. Babu, ``Capnet: Continuous
  approximation projection for {3D} point cloud reconstruction using {2D}
  supervision,'' in \emph{Proceedings of the AAAI Conference on Artificial
  Intelligence}, vol.~33, no.~01, 2019, pp. 8819--8826.

\bibitem{CNN}
Y.~LeCun, Y.~Bengio \emph{et~al.}, ``Convolutional networks for images, speech,
  and time series,'' \emph{The handbook of brain theory and neural networks},
  vol. 3361, no.~10, p. 1995, 1995.

\bibitem{DeepLearningBook}
I.~Goodfellow, Y.~Bengio, and A.~Courville, \emph{Deep learning}.\hskip 1em
  plus 0.5em minus 0.4em\relax MIT press, 2016.

\bibitem{DeepLearning}
Y.~LeCun, Y.~Bengio, and G.~Hinton, ``Deep learning,'' \emph{nature}, vol. 521,
  no. 7553, pp. 436--444, 2015.

\bibitem{GuideToCNN}
V.~Dumoulin and F.~Visin, ``A guide to convolution arithmetic for deep
  learning,'' \emph{arXiv preprint arXiv:1603.07285}, 2016.

\bibitem{Pytorch}
A.~Paszke, S.~Gross, F.~Massa, A.~Lerer, J.~Bradbury, G.~Chanan, T.~Killeen,
  Z.~Lin, N.~Gimelshein, L.~Antiga \emph{et~al.}, ``Pytorch: An imperative
  style, high-performance deep learning library,'' \emph{Advances in Neural
  Information Processing Systems}, vol.~32, pp. 8026--8037, 2019.

\bibitem{AdamOptimizer}
D.~P. Kingma and J.~Ba, ``Adam: A method for stochastic optimization,''
  \emph{arXiv preprint arXiv:1412.6980}, 2014.

\bibitem{bishop}
C.~M. Bishop, ``Pattern recognition,'' \emph{Machine learning}, vol. 128,
  no.~9, 2006.

\end{thebibliography}
}
 
\end{document}